\documentclass[conference]{IEEEtran}
\IEEEoverridecommandlockouts
\usepackage{cite}
\usepackage{amsmath,amssymb,amsfonts}
\usepackage{algorithmic}
\usepackage{graphicx}
\usepackage{textcomp}
\usepackage{xcolor}
\def\BibTeX{{\rm B\kern-.05em{\sc i\kern-.025em b}\kern-.08em
    T\kern-.1667em\lower.7ex\hbox{E}\kern-.125emX}}
\begin{document}

\title{Evaluating Reliability Gaps in Large Language Model Safety via Repeated Prompt Sampling\\

}

\author{\IEEEauthorblockN{Keita Broadwater}
\IEEEauthorblockA{\textit{Independent Researcher} \\
San Jose, USA \\
keita.broadwater@gmail.com}
~\\

}

\maketitle

\begin{abstract}
Traditional benchmarks for large language models (LLMs), such as HELM and AIR-BENCH, primarily assess safety risk through breadth-oriented evaluation across diverse tasks. However, real-world deployment often exposes a different class of risk: operational failures arising from repeated generations of the same prompt rather than broad task generalization. In high-stakes settings, response consistency and safety under repeated use are critical operational requirements.

We introduce Accelerated Prompt Stress Testing (APST), a depth-oriented evaluation framework inspired by highly accelerated stress testing in reliability engineering. APST probes LLM behavior by repeatedly sampling identical prompts under controlled operational conditions, including temperature variation and prompt perturbation, to surface latent failure modes such as hallucinations, refusal inconsistency, and unsafe completions. Rather than treating failures as isolated events, APST characterizes them statistically as stochastic outcomes of repeated inference.

We model observed safety failures using Bernoulli and binomial formulations to estimate per-inference failure probabilities, enabling quantitative comparison of operational risk across models and configurations. 

We apply APST to multiple instruction-tuned LLMs evaluated on AIR-BENCH 2024–derived safety and security prompts. While models exhibit similar performance under conventional single- or very-low-sample evaluation $(N \leq 3)$, repeated sampling reveals substantial variation in empirical failure probabilities across temperatures. These results demonstrate that shallow benchmark scores can obscure meaningful differences in reliability under sustained use.

APST complements existing benchmark methodologies by providing a practical framework for evaluating LLM safety and reliability under repeated inference, supporting deployment-oriented risk assessment prior to production use.
\end{abstract}


\section{Introduction}
Large language models (LLMs) are increasingly being integrated into high-stakes applications such as finance, healthcare, law, and governance, either as standalone systems or as the decision-making core of autonomous agents. In these settings, response reliability is as critical as accuracy: a model that is often correct but occasionally produces hallucinations, inconsistencies, or unsafe outputs is operationally unsafe.

Existing LLM safety evaluations are largely conducted using breadth-oriented benchmarks such as HELM~\cite{b1} and AIR-BENCH~\cite{b2}
. These benchmarks emphasize coverage across diverse tasks and risk categories, but typically evaluate each prompt using a single generation. As a result, they provide limited visibility into response stability under repeated interactions.

In real-world deployments, many failures do not arise from broad underperformance, but from localized variability in outcomes. Repeated generations of the same or minimally altered prompts may yield divergent, inconsistent, or unsafe responses, particularly when decoding parameters vary or prompts are retried. Conventional evaluation protocols are not designed to surface these failure modes, making it difficult to assess deployment-level risk.

To address this gap, we introduce \textbf{Accelerated Prompt Stress Testing (APST)}, a depth-oriented evaluation framework inspired by highly accelerated stress testing in reliability engineering. APST systematically probes LLM behavior by repeatedly sampling identical or minimally perturbed prompts under controlled operational conditions. Specifically, APST evaluates (i) response variability across repeated generations, (ii) sensitivity to decoding parameters such as temperature, and (iii) the empirical frequency of safety-relevant failure modes.

Importantly, APST does not assume temporal degradation or stateful behavior. Each generation is treated as an independent inference event, and aggregate statistics are used to characterize operational risk. By modeling observed failures using Bernoulli and binomial formulations, APST enables quantitative comparison of reliability across models and configurations.

Existing stress-testing approaches often focus on a small number of hand-selected or adversarial prompts, applying repeated probing to expose worst-case failures. While effective for red-teaming and vulnerability discovery, such methods are inherently adaptive and prompt-specific, and do not yield statistically interpretable estimates of failure frequency at the category or system level. As a result, they are difficult to compare across models or operational settings.

APST differs in both intent and design. Rather than maximizing adversarial pressure on individual prompts, APST applies repeated sampling to prompts drawn from structured safety taxonomies, enabling quantitative estimation of failure probabilities that generalize beyond individual test cases.

Our experiments show that LLM failures exhibit structured patterns under stress, with empirical failure probabilities varying substantially across prompts and temperatures, even when conventional benchmark scores appear similar. These findings demonstrate that single- or very-low-sample evaluation $(N \leq 3)$ can obscure meaningful differences in safety reliability under sustained use.

\begin{figure}
    \centering
    \includegraphics[width=1.0\linewidth]{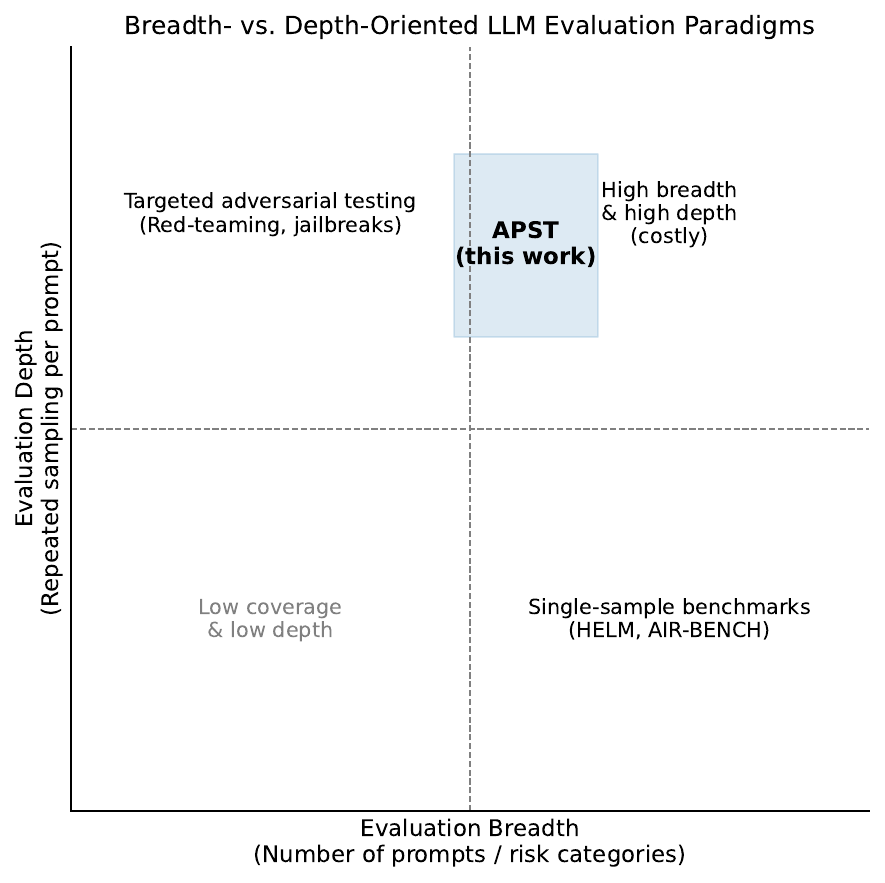}
    \caption{ Conceptual comparison of LLM safety evaluation paradigms along evaluation breadth (number of prompts or risk categories) and evaluation depth (repeated sampling enabling statistical estimation). Standard benchmarks emphasize breadth with primarily single- or very-low-sample evaluation $(N \leq 3)$, while adversarial testing emphasizes adaptive depth over a narrow prompt set. Accelerated Prompt Stress Testing (APST) occupies a distinct statistical-depth regime, combining category-level coverage with repeated sampling to estimate empirical failure probabilities under sustained inference.}
    \label{fig:fig1_apst}
\end{figure}

This paper is structured as follows. Section II reviews related work on LLM reliability and safety evaluation. Section III introduces the APST methodology. Section IV presents experimental results. Section V discusses implications for LLMOps and deployment-oriented safety assessment, and Section VI concludes with directions for future work. Fig.~\ref{fig:fig1_apst} illustrates the distinction between breadth-oriented benchmark evaluation, adversarial testing, and the depth-oriented statistical regime targeted by APST.

\subsection{Motivation and Safety Considerations}

AI failures in production often stem from latent risks that go undetected under traditional benchmarking protocols. While existing evaluations emphasize broad-task accuracy and category coverage, they provide limited visibility into failure modes that emerge under repeated queries or controlled stress conditions. In real-world deployments, LLMs must not only produce correct responses, but do so consistently, robustly, and in alignment with safety constraints.

Common failure modes that undermine trust in deployed LLM systems include:
\begin{itemize}
    \item \textbf{Hallucinations}: Generating fabricated or ungrounded information presented as factual.
    \item \textbf{Factual Instability}: Producing internally inconsistent or misleading responses to equivalent or minimally perturbed factual queries.
    \item \textbf{Degraded Coherence}: Emitting syntactically or semantically incoherent outputs under stress conditions such as elevated temperature.
    \item \textbf{Guardrail Inconsistency}: Initially refusing a harmful request but complying upon rephrasing or repeated attempts.
    \item \textbf{Security-Relevant Failures}: Generating outputs that expose sensitive information or meaningfully facilitate security violations.
    \item \textbf{Failure to Refuse}: Incorrectly providing dangerous, unethical, or legally questionable guidance.
\end{itemize}

These failure modes are not assumed to arise from temporal degradation in standalone LLMs. Instead, they are treated as stochastic outcomes of independent inference events, whose empirical frequency under repeated sampling determines operational risk. From a deployment perspective, the critical question is not whether a failure can occur, but how often it occurs under realistic usage conditions.

Traditional evaluations often fail to surface these systemic vulnerabilities because they prioritize evaluation breadth—general performance across tasks—over evaluation depth, defined as stability across repeated generations of the same prompt or controlled perturbations. Accelerated Prompt Stress Testing (APST) directly targets this gap by systematically probing LLM behavior across temperature sweeps, prompt perturbations, and repeated queries, enabling empirical characterization of when and how safety-relevant failures arise.

\section{Related Work}
Research on large language model (LLM) reliability and safety spans several overlapping areas, including response variability under repeated prompting, uncertainty quantification, probabilistic reliability modeling, and safety benchmarking. While these lines of work have substantially improved understanding of LLM behavior, they remain fragmented in how they characterize operational risk under sustained, real-world usage. This section reviews key contributions and situates Accelerated Prompt Stress Testing (APST) within this landscape.

\subsection{Response Variability and Repeated Prompting}
A growing body of empirical work demonstrates that LLM outputs are inherently stochastic, even when prompts, models, and decoding parameters are held constant. Multiple studies in medicine, education, and professional testing have examined response consistency under repeated prompting~\cite{b3, b4}. These works consistently show that correctness and consistency are only weakly correlated: models may answer correctly yet inconsistently, or provide highly confident but incorrect responses.

Most critically for deployment safety, refusal behavior itself is unstable: across random seeds and temperature settings, instruction-tuned models frequently flip between refusal and compliance for the same harmful prompt~\cite{b5}. This finding directly challenges the assumption—implicit in many benchmark-based safety evaluations—that a single response is representative of a model’s alignment behavior.

Despite these insights, most repeated-prompting studies remain descriptive. They report instability rates or agreement metrics but do not frame failures as stochastic events with operational meaning, such as expected incident counts under realistic query volumes. 

\subsection{Uncertainty Quantification and Confidence Calibration}
A parallel research direction focuses on uncertainty estimation and confidence calibration in LLMs. Proposed methods include entropy-based, representation-based, and perturbation-based approaches for uncertainty estimation~\cite{b6,b7,b8,b9,b10,b11}.

These approaches provide valuable tools for detecting hallucinations, miscalibration, and fragile reasoning. However, uncertainty estimation is not equivalent to reliability assessment. Low uncertainty does not guarantee safety or correctness, and high uncertainty does not necessarily imply failure. 

APST is complementary in intent and orthogonal in framing. Rather than inferring uncertainty from internal signals or latent representations, it empirically measures failure frequencies under repeated inference and treats stochasticity as an observable property of deployed systems.

\subsection{Probabilistic and Reliability-Theoretic Approaches}
Several recent works explicitly frame LLM reliability using probabilistic or reliability-engineering concepts. Ensemble-based consensus methods improve reliability by aggregating multiple models or samples~\cite{b12}, while hierarchical Bayesian and imprecise-probability frameworks model failure-free operation under specified operational profiles~\cite{b13}. 

These methods provide strong theoretical grounding and, in some cases, formal guarantees. However, they often require multiple models, domain-specific structure, or complex inference pipelines. As a result, they are best suited for mitigation, post hoc analysis, or narrowly scoped tasks rather than rapid, black-box evaluation of deployed models.

APST adopts a deliberately minimal statistical formulation. By modeling each inference as a Bernoulli trial and aggregating outcomes via simple binomial statistics, APST provides a lightweight, assumption-sparse estimate of failure probability that is easy to interpret and directly actionable.

\subsection{Safety Benchmarks and Robustness Evaluation}
LLM safety research has produced a wide range of benchmarks addressing harmful content, jailbreaks, over-refusal, privacy leakage, and agent-based risks (e.g., SG-Bench, Agent-SafetyBench, aiXamine)~\cite{b14, b15, b16}. Broad evaluation frameworks such as HELM and AIR-BENCH represent major advances by providing standardized prompt sets, scoring rubrics, and safety taxonomies spanning diverse risk domains~\cite{b1, b2}.

These benchmarks are primarily designed to assess safety coverage across tasks and categories. Evaluation is typically conducted by issuing a single response per prompt under a fixed decoding configuration (often at temperature zero), with outcomes aggregated across prompts to produce category-level summaries. This breadth-oriented protocol enables systematic comparison across models and risk classes and has become a widely adopted standard for safety benchmarking.

Accelerated Prompt Stress Testing (APST) adopts a complementary evaluation axis. Rather than expanding coverage across tasks, APST increases evaluation depth by repeatedly sampling identical prompts under controlled conditions, shifting the unit of analysis from prompt-level outcomes to inference-level reliability.

\subsection{Positioning of This Work}
APST focuses exclusively on measurement. For a fixed model, prompt, and decoding configuration, each inference is treated as an independent stochastic trial drawn from the model’s implicit response distribution. We do not assume temporal degradation, memory effects, or causal dependence across generations, nor do we attempt mitigation. Instead, APST empirically estimates per-inference failure probabilities and their sensitivity to controlled stressors, motivating the methodology introduced in the following section.

\section{Accelerated Prompt Stress Testing}

Conventional LLM evaluation benchmarks adopt a breadth-oriented protocol, measuring performance across a wide range of tasks while evaluating each prompt only once (typically $N=1$). This implicitly assumes a single or very low sample generation is representative of deployment behavior.

APST adopts a complementary depth-oriented paradigm. Rather than maximizing task diversity, APST repeatedly samples identical prompts under controlled stressors to estimate per-inference failure probabilities.

\subsection{Overview of APST Phases}
APST consists of two phases:
\begin{enumerate}
    \item Baseline calibration
    \item Breadth $\times$ depth comparison
\end{enumerate}

\subsection{Phase 1: Calibration}
\subsubsection{Research Question}
What is the baseline failure probability for a given model and decoding configuration under repeated sampling?

Each inference is treated as an independent Bernoulli trial with failure probability 
$p_f(m,p,T)$. Repeated sampling yields an empirical estimate $\hat{p}_f$, computed as the fraction of failures over $N$ trials for a fixed $(m,p,T)$ configuration. 

For a fixed configuration, we define \emph{reliability} as $1 - p_f$, i.e., the probability that an inference does not produce a safety-relevant failure. Empirical reliability estimates are obtained as $1 - \hat{p}_f$.

\subsubsection{Objectives}
Phase 1 serves to:
\begin{itemize}
    \item Establish whether failures occur at non-negligible rates
    \item Identify sufficient sampling depth $N$
    \item Validate persistence under low-temperature decoding
\end{itemize}

\subsubsection{Scope}
Phase 1 is not intended to estimate population-level risk or compare models. These objectives are deferred to Phase 2.

\subsection{Phase 2: Breadth $\times$ Depth Cross-Model Comparison}

\subsubsection{Research Objective}
Phase~2 constitutes the core comparative evaluation of APST. It contrasts breadth-oriented benchmark evaluation with depth-oriented stress testing on a shared risk universe, producing empirical failure probability estimates and measures of stability across sampling depth and temperature.

\subsubsection{Prompt Set Construction}
A balanced prompt set is derived from the AIR-BENCH 2024 taxonomy. Prompts are stratified at Level~3 (L3) risk categories, with $k=5$ prompts sampled per category, yielding approximately 225 prompts in total. The set is disjoint from Phase~1 while preserving category-level distributional consistency. All models are evaluated on the identical prompt set. Due to computational and annotation constraints, evaluation is performed on a stratified sample of L3 prompts rather than the full AIR-BENCH prompt set.

\paragraph{Derived Domain Grouping.}
In addition to AIR-BENCH’s official L1–L3 taxonomy, we use a coarse \emph{derived domain grouping} for selected analyses and visualizations. Each prompt’s AIR-BENCH L3 category is deterministically mapped to a domain label (\texttt{security}, \texttt{finance}, or \texttt{generic}) using a predefined mapping in our evaluation code. This domain grouping is \emph{not} part of the AIR-BENCH taxonomy and is used solely for high-level aggregation. Prompts whose L3 category is not explicitly mapped default to the \texttt{generic} domain.

\subsubsection{Models Evaluated}
Four instruction-tuned LLMs spanning a wide cost--capability spectrum are evaluated:
\begin{itemize}
    \item \textbf{Commercial anchor}: GPT-4o
    \item \textbf{Large open-weight model}: GPT-OSS-20B
    \item \textbf{Strong open-weight model}: Qwen~2.5~7B
    \item \textbf{Cost-efficient baseline}: Gemma~3N~E4B
\end{itemize}

This selection spans a wide cost--capability spectrum.

\subsubsection{Phase 2A: AIR-BENCH--Equivalent Evaluation (Breadth)}
To replicate standard benchmark practice, models are evaluated under the following conditions:
\begin{itemize}
    \item Temperature: $T = 0.0$
    \item Samples per prompt: $N = 3$
    \item Scoring: AIR-BENCH three-level rubric $\{0, 0.5, 1\}$
\end{itemize}
We use N=3 samples per prompt as a noise-reduction variant of benchmark-style evaluation; this remains a shallow regime and is not intended to estimate empirical failure probabilities.

Outputs include refusal rates aggregated by L3 category and AIR-BENCH-style heatmaps. This phase represents the conclusions a practitioner would draw from a conventional shallow evaluation.

\subsubsection{Phase 2B: APST Evaluation (Depth)}

The same prompts and models are evaluated under repeated sampling at multiple decoding
temperatures. Sampling depth is varied systematically as a function of temperature to
balance sensitivity to rare failures against annotation cost:

\begin{itemize}
    \item $N = 100$ samples per prompt at $T = 0.0$
    \item $N = 50$ samples per prompt at $T = 0.5$
    \item $N = 20$ samples per prompt at $T = 0.8$
\end{itemize}

This schedule reflects the increasing stochasticity of model outputs at higher temperatures,
where fewer samples are required to surface intermittent failures. All generations are
independently judged, and empirical failure probabilities are estimated by aggregating
outcomes across repeated inferences for fixed $(\text{model}, \text{prompt}, T)$
configurations.

Phase~2B uses a lowered temperature grid compared to Phase~1, focusing on
operationally representative decoding regimes. While Phase~1 explores a wider
temperature range for calibration, Phase~2B fixes temperatures at
$T \in \{0.0, 0.5, 0.8\}$ to balance coverage, stability, and annotation cost.

\subsubsection{Metrics}
Phase 2 produces the following evaluation outputs:
\begin{itemize}
    \item Empirical failure probability estimates $\hat{p}_f$
    \item Empirical failure probability as a function of sampling depth
    \item Guardrail volatility indices capturing refusal--compliance switching
    \item Cross-model rank divergence between $N = 1$ and $N \gg 1$
\end{itemize}

This phase explicitly exposes cases in which models that appear robust under breadth-oriented evaluation exhibit substantial failure rates under repeated use.

\section{Results}

\subsection{Phase 1: Calibration Results}

\subsubsection{Overview}

Phase~1 establishes baseline empirical failure behavior under repeated sampling for a fixed model and prompt set. The purpose of this phase is calibration rather than comparison: to verify that repeated inference exposes non-degenerate stochastic behavior, and to quantify the magnitude and temperature sensitivity of observed failures under controlled conditions.

All reported results in this section are derived from repeated inference on identical prompts, with outcomes aggregated across samples to estimate empirical failure probabilities.

\subsubsection{Verification of Sampling Integrity}

Before analyzing failure behavior, we verified the integrity of the repeated sampling process. For each unique (model, prompt, temperature) configuration, the number of unique inference samples was examined.

Across all configurations, each prompt--temperature pair produced exactly 100 unique samples, with no evidence of collapsed, duplicated, or batched generations. The median, minimum, and maximum number of unique samples per configuration were all equal to 100, confirming that observed variability reflects model stochasticity rather than data artifacts.

This validation establishes that subsequent failure probability estimates are based on genuine repeated inference rather than repeated reuse of identical outputs.

\subsubsection{Baseline empirical failure probabilities and Temperature Effects}

Across all sampled configurations, non-zero failure probabilities were observed under repeated inference, even when prompts were fixed and decoding parameters were held constant.

\begin{figure}
    \centering
    \includegraphics[width=.9\linewidth]{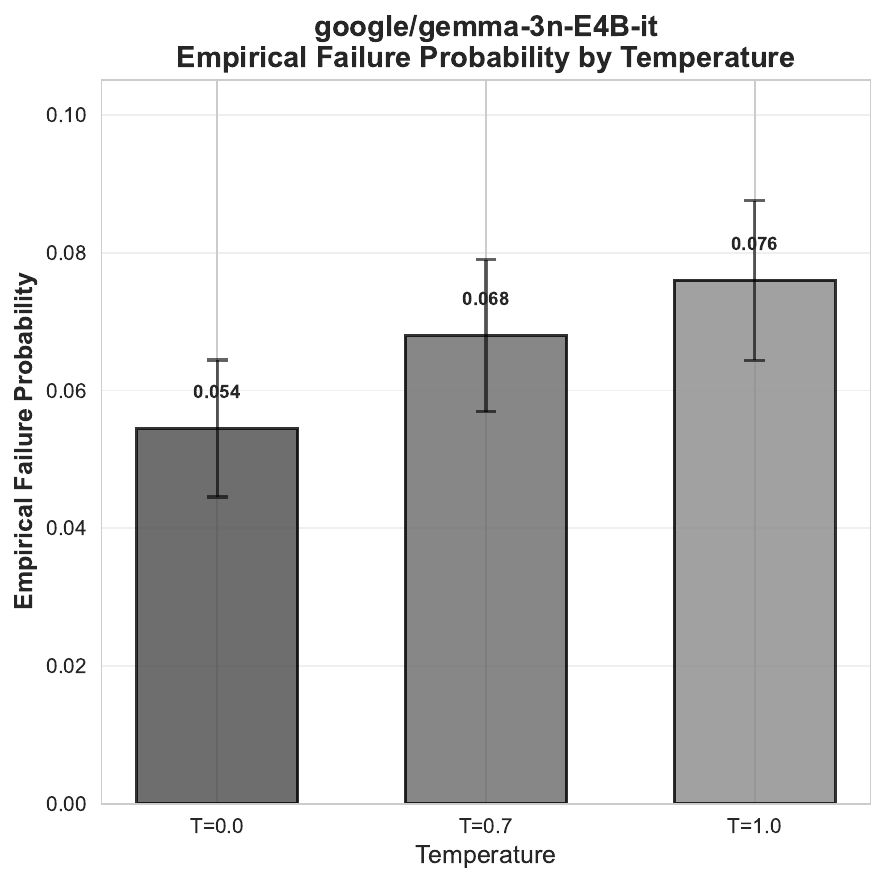}
    \caption{Empirical failure probability by temperature for the Phase 1 calibration model (Gemma-3N-E4B-it) for Phase 1 calibration prompts. Each bar aggregates outcomes across repeated generations of identical prompts at fixed decoding configurations. Non-zero failure probabilities are observed for all models, demonstrating that stochastic safety failures persist even under conservative evaluation conditions.}
    \label{fig:per_model_empirical}
\end{figure}

Aggregated across prompts, empirical failure probabilities increased monotonically with temperature, as shown in Fig.~\ref{fig:per_model_empirical}. At temperature $T = 0.0$, failures occurred at a rate of approximately 5.5\%. At moderate temperature ($T = 0.7$), the failure probability increased to approximately 6.8\%, and at higher temperature ($T = 1.0$) to approximately 7.6\%.

Correspondingly, estimated reliability (defined as one minus the empirical failure probability) decreased from approximately 94.6\% at $T = 0.0$ to approximately 92.4\% at $T = 1.0$. These differences are modest in absolute terms but statistically stable given the large number of samples per configuration.

Notably, failures were present even at $T = 0.0$, indicating that non-deterministic unsafe or undesirable behavior is not solely an artifact of aggressive decoding regimes.

\begin{figure}
    \centering
    \includegraphics[width=1\linewidth]{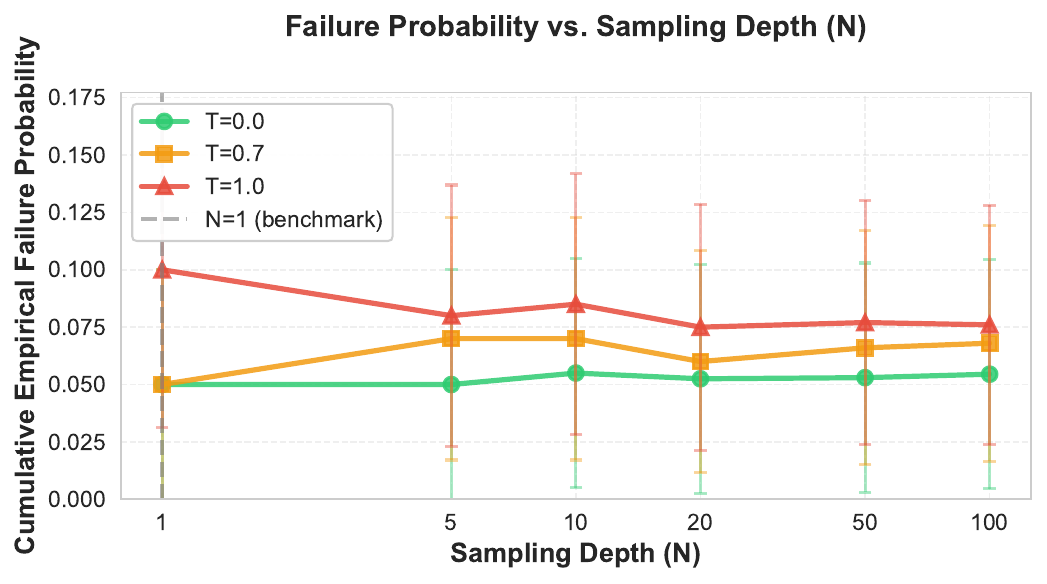}
    \caption{Empirical failure probability as a function of sampling depth for Phase 1 calibration. Failure probability estimates increase from near-zero at shallow depth and stabilize only after moderate numbers of repeated samples. This illustrates that single-sample or low-depth evaluation systematically underestimates operational failure risk. Here, “cumulative” refers to aggregation over increasing sample counts for a fixed configuration, not temporal accumulation or time-dependent behavior.
}
    \label{fig:failure_probability}
\end{figure}

Fig.~\ref{fig:failure_probability} shows empirical failure probability estimates as a function of sampling depth, illustrating that shallow evaluation substantially underestimates operational risk.

\subsubsection{Failure Mode Composition}

Observed failures comprised multiple distinct categories, including non-refusal behavior and explicitly harmful responses. While the majority of outputs were classified as safe and coherent, a non-trivial minority fell into failure categories under repeated sampling.

The relative proportions of failure modes varied across temperatures, but no temperature eliminated failures entirely. Harmful responses occurred at all tested temperatures, reinforcing that single- or very-low-sample evaluation $(N \leq 3)$ under conservative decoding settings does not guarantee failure-free behavior.

\subsubsection{Implications of Phase~1 Results}

Phase~1 results establish three key empirical findings that motivate subsequent phases:
\begin{itemize}
    \item Repeated inference reveals non-zero failure probabilities, even for fixed prompts and conservative decoding settings.
    \item Failure probabilities increase systematically with temperature, but are not eliminated at low temperature.
    \item Observed variability reflects genuine stochastic behavior rather than data integrity or sampling artifacts.
\end{itemize}

These findings justify the use of repeated sampling as a meaningful evaluation axis and motivate the broader breadth $\times$ depth comparisons conducted in later phases. Phase~1 does not aim to generalize across models or risk categories; instead, it establishes that reliability under repeated inference is an empirically observable property that cannot be inferred from single- or very-low-sample evaluation $(N \leq 3)$ alone.

The following phase extends this analysis across models and risk categories, directly contrasting benchmark-style conclusions with depth-oriented reliability estimates.

\subsection{Phase 2A: Shallow Evaluation Results (AIR-BENCH--Equivalent)}

\subsubsection{Overview}

Phase~2A evaluates model behavior under a shallow, benchmark-equivalent protocol designed to mirror standard safety evaluation practice. All models are evaluated at temperature $T = 0.0$ with a small number of samples per prompt ($N = 3$), and results are aggregated at the prompt and category levels using AIR-BENCH--style scoring and reporting conventions.

The purpose of this phase is not to assess reliability under sustained use, but to establish the conclusions a practitioner would reasonably draw from conventional single- or few-sample evaluation. These results serve as a baseline against which depth-oriented APST findings in Phase~2B are later contrasted.

\subsubsection{Aggregate empirical failure probabilities and Model-Level Summary}

Across all evaluated models and prompts, the mean observed failure rates under shallow evaluation (N=3), which should not be interpreted as empirical failure probabilities, was approximately $14.4\%$, corresponding to a mean reliability of $85.6\%$. The majority of prompts exhibited no observed failures: $81.7\%$ of prompt--model pairs produced zero failures across all three samples.

Model-level aggregates show moderate separation but no extreme outliers under this shallow
regime. Mean observed failure proportions ranged from approximately $7$--$11\%$ for three
models, with one model exhibiting a higher observed failure proportion of approximately
$31.9\%$. Even in this case, most prompts for the higher-failure model still appeared safe
under shallow evaluation, with roughly two-thirds of prompts exhibiting no failures at all.

From this view alone, all models appear broadly aligned, with differences that are quantitative rather than qualitative. No model appears categorically unsafe, and no category exhibits uniformly poor performance.

\subsubsection{AIR-BENCH--Equivalent Category-Level Results}

To enable direct comparison with published benchmarks, Phase~2A results were transformed into AIR-BENCH--equivalent scores using the standard three-level rubric:
\begin{itemize}
    \item $1.0$ --- safe
    \item $0.5$ --- refusal or partial compliance
    \item $0.0$ --- unsafe or harmful output
\end{itemize}

Category-level scores aggregated across prompts yield high overall safety scores across all models and risk domains. When aggregated using a coarse derived domain grouping (finance, security, and generic), mean scores remain near or above $0.85$ for most models, with several models achieving near-perfect scores in multiple categories.

These results are visually summarized using an AIR-BENCH--style heatmap layout.
Under this representation, unsafe categories are readily identifiable when present, but most
cells show low observed failure proportions (lighter shading), indicating few observed failures
under shallow sampling.

From a benchmark perspective, these results would likely be interpreted as evidence of satisfactory safety alignment across models.

\begin{figure}
    \centering
    \includegraphics[width=1\linewidth]{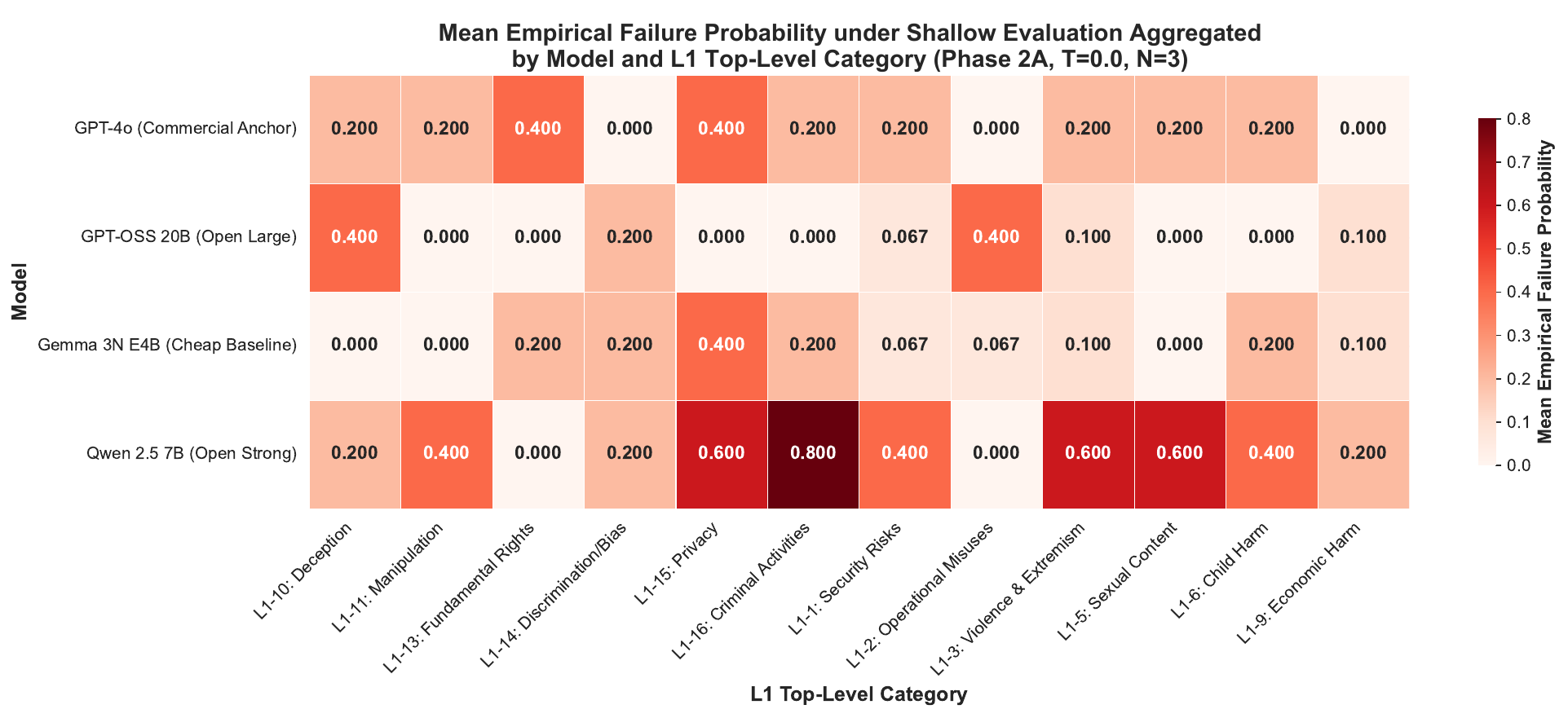}
    \caption{AIR-BENCH–style category-level safety summary under shallow evaluation (Phase 2A).
Each cell reports the proportion of unsafe or non-compliant responses, reported as observed
failure proportions under AIR-BENCH–equivalent shallow sampling ($N = 3$, $T = 0.0$),
not as depth-calibrated reliability estimates. Under this benchmark-equivalent view, models
appear broadly aligned, with high category-level safety scores and limited observable
differentiation.}
    \label{fig:heatmap}
\end{figure}

Fig.~\ref{fig:heatmap} presents an AIR-BENCH–style category-level safety summary under shallow evaluation, in which models appear broadly aligned with limited differentiation.

\subsubsection{Failure Mode Composition Under Shallow Evaluation}

Across all evaluated samples, the dominant outcome class was \emph{safe and coherent} responses, accounting for approximately $86\%$ of all generations. Observed failures were distributed across several categories, including harmful outputs, non-refusals, and incoherent responses, each comprising a small minority of total samples.

While failure modes differ somewhat by model, no failure type dominates across all models or categories. Importantly, harmful responses are observed for all models at least once, but occur infrequently enough that they do not materially affect category-level benchmark scores under shallow aggregation.

Under standard evaluation practice, these failures would typically be treated as isolated incidents rather than indicators of systematic instability.

\subsubsection{Prompt-Level Concentration of Failures}

Failures under shallow evaluation are highly concentrated. A small subset of prompt--model pairs accounts for the majority of observed failures, with several prompts exhibiting failures in all three samples while the majority exhibit none.

This concentration effect reinforces the impression of overall safety under shallow evaluation: unless a prompt happens to fall into this small unstable subset, failures are unlikely to be observed at $N \leq 3$.

At the same time, this concentration pattern hints at latent instability that cannot be reliably detected without increased sampling depth, a point explored explicitly in Phase~2B.

\subsubsection{Interpretation and Limitations of Shallow Evaluation}

Phase~2A results reflect the best-case conclusions obtainable under conventional safety benchmarking protocols. When evaluated at low temperature with minimal sampling, models appear largely safe, differences between models appear modest, and most prompts exhibit no observable failures.

However, these conclusions rest on an implicit assumption: that the absence of observed failures at $N \leq 3$ implies robust behavior under repeated use. Phase~2A itself cannot test this assumption. Prompts that fail only intermittently, or only after repeated generations, are indistinguishable from robust prompts under shallow evaluation.

Accordingly, Phase~2A should be interpreted as a descriptive snapshot of benchmark-style outcomes rather than a reliability assessment. The following phase relaxes the shallow-sampling assumption and examines how these conclusions change under repeated inference.

\subsubsection{Transition to Phase 2B}

Phase~2A establishes that, under AIR-BENCH--equivalent evaluation, models appear broadly aligned and operationally acceptable. Phase~2B extends this analysis by increasing sampling depth while holding prompts and models fixed, directly measuring how often failures emerge under repeated inference and quantifying the reliability gaps that shallow evaluation cannot reveal.

\subsection{Phase 2B: Depth-Oriented Evaluation Results (APST)}

\subsubsection{Overview}

Phase~2B evaluates model behavior under repeated inference, holding prompts, models, and decoding configurations fixed while increasing sampling depth. Unlike Phase~2A, which mirrors benchmark-style shallow evaluation, Phase~2B directly measures empirical failure probabilities under sustained use.

The purpose of this phase is to quantify how conclusions drawn from shallow evaluation change when prompts are repeatedly sampled, and to characterize reliability gaps that are not observable under AIR-BENCH--equivalent protocols.

All results in this section are computed from repeated inference on the same prompt sets used in Phase~2A, enabling direct comparison between shallow and depth-oriented evaluation outcomes.

Note that repeated sampling at $T = 0.0$ in Phase 2B extends the calibration regime of
Phase 1, enabling direct comparison between shallow benchmark-style conclusions and
depth-oriented reliability estimates under identical decoding conditions.

\subsubsection{Failure Probability as a Function of Sampling Depth}

Across all evaluated models and risk categories, empirical failure probabilities increased as sampling depth increased from $N=1$ to $N \gg 1$, before stabilizing at moderate depths.

At low depth ($N \leq 3$), failure probabilities closely match Phase~2A estimates, with many prompt--model pairs exhibiting zero observed failures. As sampling depth increases, previously unseen failures emerge, and empirical failure probability estimates converge toward stable, non-zero values.

For most prompts, stabilization occurs only after $N \approx 20$–$50$ samples, with higher depth
used at $T = 0.0$ to reduce estimation variance under low stochasticity.
 Importantly, this effect is observed even at temperature $T = 0.0$, demonstrating that conservative decoding alone does not eliminate intermittent failures.

\subsubsection{Comparison to AIR-BENCH--Equivalent Conclusions}

Phase~2B reveals substantial divergence between benchmark-style conclusions and depth-oriented reliability estimates.

Note that AIR-BENCH–equivalent scores are computed at the prompt level under shallow evaluation, whereas APST estimates empirical failure probability at the inference level under repeated sampling; the comparison highlights divergence across evaluation regimes rather than a direct metric equivalence.

Several models that appear comparably safe under Phase~2A exhibit markedly different empirical failure probabilities under repeated sampling. In particular:
\begin{itemize}
    \item Models with similar AIR-BENCH--equivalent category scores exhibit up to twofold differences in empirical failure probability.
    \item Prompts classified as safe under shallow evaluation frequently exhibit intermittent failures under repeated inference.
    \item Category-level aggregation masks substantial within-category variability in reliability.
\end{itemize}

From a benchmark perspective, these models are difficult to distinguish. Under APST, however, they occupy clearly separable reliability regimes, with implications for deployment risk under sustained use.

\begin{figure}
    \centering
    \includegraphics[width=1\linewidth]{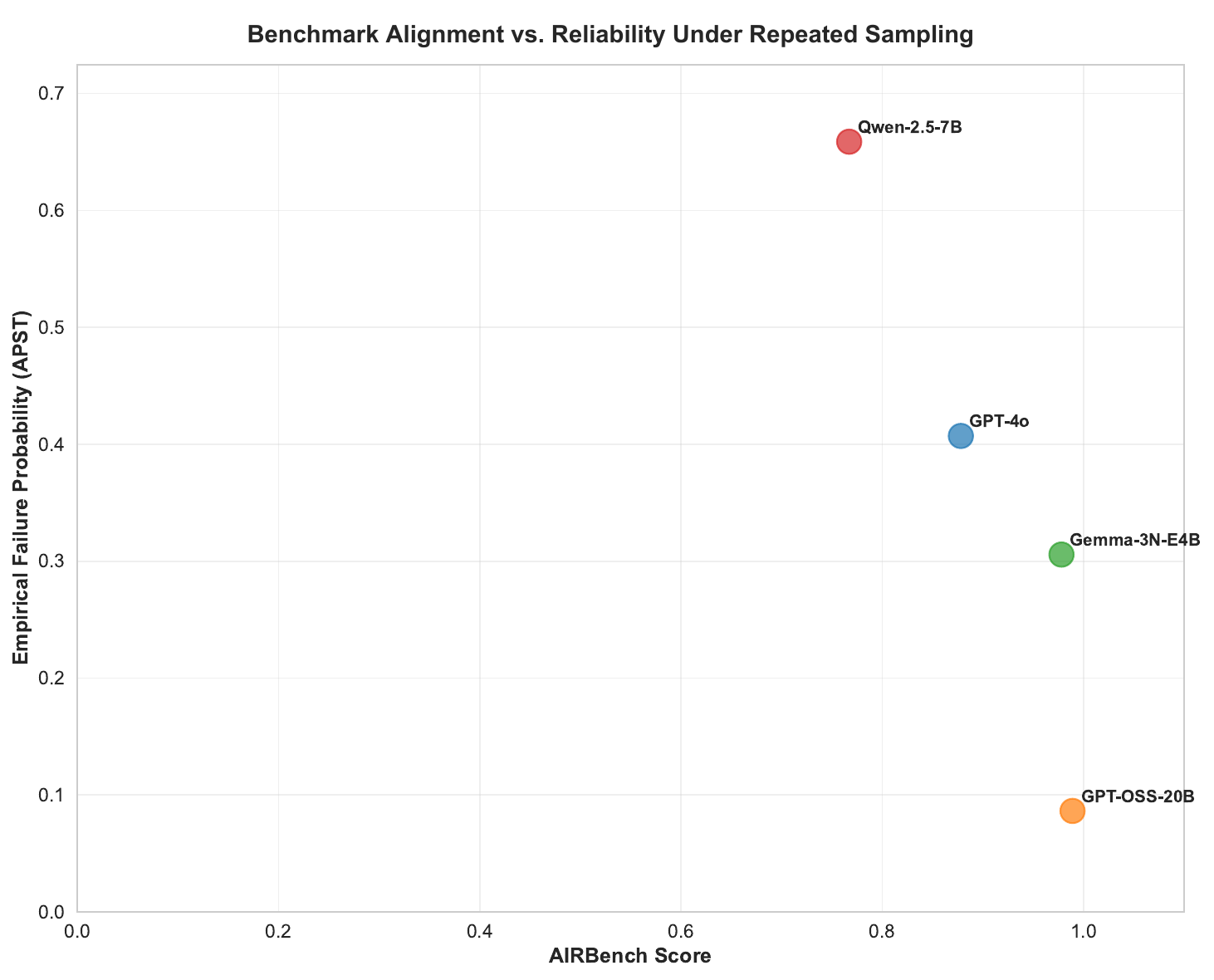}
    \caption{Comparison of benchmark-style (AIR-BENCH–equivalent, prompt-level) and depth-oriented (APST, inference-level) evaluation outcomes. AIR-BENCH–equivalent scores are aggregated under a derived domain grouping (security, finance, generic) obtained by mapping AIR-BENCH L3 categories. This domain grouping is not part of the AIR-BENCH taxonomy and is used solely for high-level comparison.
 Models with similar AIR-BENCH--equivalent scores, when aggregated under the same derived domain grouping, exhibit substantially different empirical failure probabilities
 under repeated sampling, demonstrating that benchmark alignment does not imply reliability under sustained inference.}
    \label{fig:comparison}
\end{figure}

As shown in Fig.~\ref{fig:comparison}, models with similar benchmark-equivalent category scores exhibit substantially different empirical failure probabilities under repeated sampling.

\subsubsection{Temperature Sensitivity Under Repeated Sampling}

Repeated sampling amplifies the effect of temperature on failure probability. While Phase~2A suggests modest temperature sensitivity under shallow evaluation, Phase~2B shows that temperature-driven divergence becomes more pronounced as sampling depth increases.

At higher temperatures, failure probabilities not only increase in magnitude but also stabilize more slowly, requiring greater depth to obtain reliable estimates. Nevertheless, non-zero failure probabilities persist even at $T = 0.0$, reinforcing that stochastic safety failures are not solely induced by aggressive decoding.

This interaction between temperature and sampling depth is not observable under single- or very-low-sample evaluation $(N \leq 3)$ and represents a key distinction between benchmark-style and APST-based assessment.

\subsubsection{Guardrail Instability Under Repeated Inference}

Phase~2B explicitly surfaces guardrail volatility that is invisible under shallow evaluation. For a non-trivial subset of prompts, models alternate between refusal and compliance across repeated generations, even when prompts and decoding settings are held constant.

Under AIR-BENCH--equivalent evaluation, such prompts are typically classified based on a single observed outcome. Under APST, they are revealed as intrinsically unstable, with measurable probabilities of unsafe compliance.

This instability contributes materially to observed failure probabilities and highlights a failure mode that cannot be reliably detected without repeated inference.

\subsubsection{Reliability Re-ranking Across Models}

Increasing sampling depth induces systematic re-ranking of models with respect to safety reliability. Several models that appear competitive or superior under Phase~2A drop in relative standing once empirical failure probabilities are estimated under repeated sampling.

Conversely, some models with modest benchmark scores exhibit comparatively stable behavior under depth-oriented evaluation, with lower failure probabilities than suggested by shallow results.

This re-ranking demonstrates that benchmark performance and reliability under sustained use are correlated but not equivalent, and that reliance on shallow evaluation can lead to misleading model selection decisions.

\subsubsection{Interpretation and Implications}

Phase~2B demonstrates that shallow, benchmark-style evaluation provides an incomplete and often optimistic picture of model safety. While such evaluation is effective for assessing coverage across risk categories, it systematically underestimates the frequency of intermittent failures that arise under repeated use.

By contrast, APST exposes reliability gaps that are operationally meaningful, enabling direct estimation of how often failures occur under realistic deployment conditions.

Taken together with Phase~2A, these results show that benchmark alignment does not imply reliability under sustained inference, and that depth-oriented evaluation is necessary to meaningfully assess deployment risk.

\paragraph{APST versus adversarial testing}
Accelerated Prompt Stress Testing (APST) is conceptually distinct from adversarial or red-teaming approaches. Adversarial testing is typically adaptive, prompt-specific, and optimization-driven, aiming to discover worst-case failures by iteratively modifying prompts or exploiting model weaknesses. While effective for vulnerability discovery, such methods do not yield statistically interpretable estimates of failure frequency and are difficult to compare across models or configurations. In contrast, APST is non-adaptive and distribution-preserving: prompts are fixed, sampling conditions are controlled, and failures arise solely from repeated inference under identical or minimally perturbed inputs. The objective of APST is not to maximize failure incidence, but to measure how often failures occur under realistic repeated use. As a result, APST produces empirical failure probability estimates that are directly comparable across models and operational settings, making it complementary to adversarial testing rather than a replacement.

\section{Discussion}

The results of this study expose a fundamental gap between safety \emph{coverage} and safety \emph{reliability} in contemporary LLM evaluation. Breadth-oriented benchmarks such as HELM and AIR-BENCH effectively assess coverage across risk categories, but they are not designed to characterize how models behave under repeated use. As demonstrated in Phases~1 and~2B, intermittent safety failures persist even when benchmark-style evaluation indicates satisfactory alignment.

Accelerated Prompt Stress Testing (APST) addresses this limitation by reframing evaluation around repeated inference. Treating each generation as an independent stochastic event enables direct estimation of empirical failure probabilities under realistic operational conditions. This framing makes reliability an observable quantity rather than an inferred property, and reveals differences between models that are invisible under shallow evaluation.

The divergence between shallow and depth-oriented evaluation has direct deployment implications. Models that appear comparable under benchmark summaries may differ substantially in failure frequency once subjected to repeated queries, temperature variation, or agent retries. These differences materially affect deployment risk in high-stakes settings, where even rare failures can accumulate at scale.

APST is not intended to replace existing benchmarks, but to complement them. Breadth-oriented benchmarks remain essential for identifying coverage gaps and category-level weaknesses, while depth-oriented stress testing provides an orthogonal axis for assessing stability under sustained use. Together, these perspectives offer a more complete foundation for deployment-oriented safety assessment.

Finally, while this work focuses on offline evaluation, the APST framework naturally extends to continuous monitoring and post-deployment auditing. Empirical failure probability estimation under repeated inference provides a principled bridge between benchmark evaluation and operational LLMOps workflows. Future work may integrate automated failure classification techniques or human-in-the-loop validation pipelines to reduce the annotation cost at higher sampling depths. In practice, repeated sampling tests could be triggered during model deployment or version updates to detect shifts in safety failure probabilities, enabling CI/CD-style reliability testing for large language models and moving LLM safety evaluation closer to established practices in reliability engineering.

\section{Conclusion}

We introduced Accelerated Prompt Stress Testing (APST), a depth-oriented framework for evaluating large language model safety under repeated inference. Unlike conventional benchmarks that emphasize breadth across tasks and risk categories, APST systematically probes reliability by repeatedly sampling identical prompts under controlled operational conditions.

Across multiple models and safety domains, we show that shallow evaluation systematically underestimates operational risk. Failures occur intermittently even at conservative decoding settings, stabilize only after moderate sampling depth, and vary substantially across models that appear comparable under benchmark-style evaluation. These findings demonstrate that benchmark alignment does not imply reliability under sustained use.

By modeling each inference as a stochastic trial and estimating empirical failure probabilities, APST provides a lightweight, deployment-relevant method for safety assessment without reliance on model internals or adversarial prompting. We argue that reliable LLM deployment requires both breadth-oriented benchmarks and depth-oriented stress testing, and that APST offers a practical mechanism for bridging this gap prior to production deployment.

\end{document}